\documentclass[letterpaper]{article} % DO NOT CHANGE THIS
\usepackage[]{aaai25}
\nocopyright 
\usepackage{booktabs}
\usepackage{multirow}
 
\usepackage{times}  % DO NOT CHANGE THIS
\usepackage{helvet}  % DO NOT CHANGE THIS
\usepackage{courier}  % DO NOT CHANGE THIS
\usepackage[hyphens]{url}  % DO NOT CHANGE THIS
\usepackage{graphicx} % DO NOT CHANGE THIS
\usepackage{natbib}  % DO NOT CHANGE THIS AND DO NOT ADD ANY OPTIONS TO IT
\usepackage{caption} % DO NOT CHANGE THIS AND DO NOT ADD ANY OPTIONS TO IT
\usepackage{algorithm}
\usepackage{algorithmic}

\usepackage{newfloat}
\usepackage{listings}
\DeclareCaptionStyle{ruled}{labelfont=normalfont,labelsep=colon,strut=off} % DO NOT CHANGE THIS
\floatstyle{ruled}
\newfloat{listing}{tb}{lst}{}
\floatname{listing}{Listing}
\title{ONRW: Optimizing inversion noise for high-quality and robust watermark}
\author {
    Xuan Ding\textsuperscript{\rm 1,\rm 2}, \ 
    Xiu Yan\textsuperscript{\rm 3}, \ 
    Chuanlong Xie\textsuperscript{\rm 4}, \ 
    Yao Zhu\textsuperscript{\rm 5}\footnotemark[1] 
}
\affiliations {
     \textsuperscript{\rm 1}Shenzhen Future Network of Intelligence Institute and Guangdong Provincial Key Laboratory of Future Networks of
Intelligence, The Chinese University of Hong Kong (Shenzhen), \\
     \textsuperscript{\rm 2}School of Science and Engineering, The Chinese University of Hong Kong (Shenzhen), \\
     \textsuperscript{\rm 3}Meituan Group, 
     \textsuperscript{\rm 4}Beijing Normal University, 
     \textsuperscript{\rm 5}Zhejiang University
 
    202322011119@mail.bnu.edu.cn, ee\_zhuy@zju.edu.cn
}

\begin{document}

\maketitle

\begin{abstract}
Watermarking methods have always been effective means of protecting intellectual property, yet they face significant challenges. Although existing deep learning-based watermarking systems can hide watermarks in images with minimal impact on image quality, they often lack robustness when encountering image corruptions during transmission, which undermines their practical application value. To this end, we propose a high-quality and robust watermark framework based on the diffusion model. Our method first converts the clean image into inversion noise through a null-text optimization process, and after optimizing the inversion noise in the latent space, it produces a high-quality watermarked image through an iterative denoising process of the diffusion model. The iterative denoising process serves as a powerful purification mechanism, ensuring both the visual quality of the watermarked image and enhancing the robustness of the watermark against various corruptions. To prevent the optimizing of inversion noise from distorting the original semantics of the image, we specifically introduced self-attention constraints and pseudo-mask strategies. Extensive experimental results demonstrate the superior performance of our method against various image corruptions. In particular, our method outperforms the stable signature method by an average of 10\% across 12 different image transformations on COCO datasets. Our codes are available at https://github.com/920927/ONRW.

\end{abstract}

\section{Introduction}\label{sec1}

The rapid advancements in generative modeling technologies have led to an exponential increase in AI-generated content. For instance, image editing tools based on diffusion models, such as ControlNet \citep{zhang2023adding} and DreamBooth \citep{ruiz2023dreambooth}, are becoming increasingly widespread. However, the extensive application of generative AI has also raised concerns regarding the authenticity and originality of content. The difficulty in accurately verifying the origin of content presents challenges in combating risks such as deepfakes, identity impersonation, and copyright infringement \citep{brundage2018malicious,bautista2024ethical, zhu2023information}. Digital watermark is a technique used to embed specific information into images. This method enables social media platforms, new agencies, and dissemination platforms to authenticate images, thereby effectively reducing potential risks associated with generative contents \citep{singh2013survey}.

\begin{figure}[h]
\centering
\includegraphics[width=1.0\columnwidth]{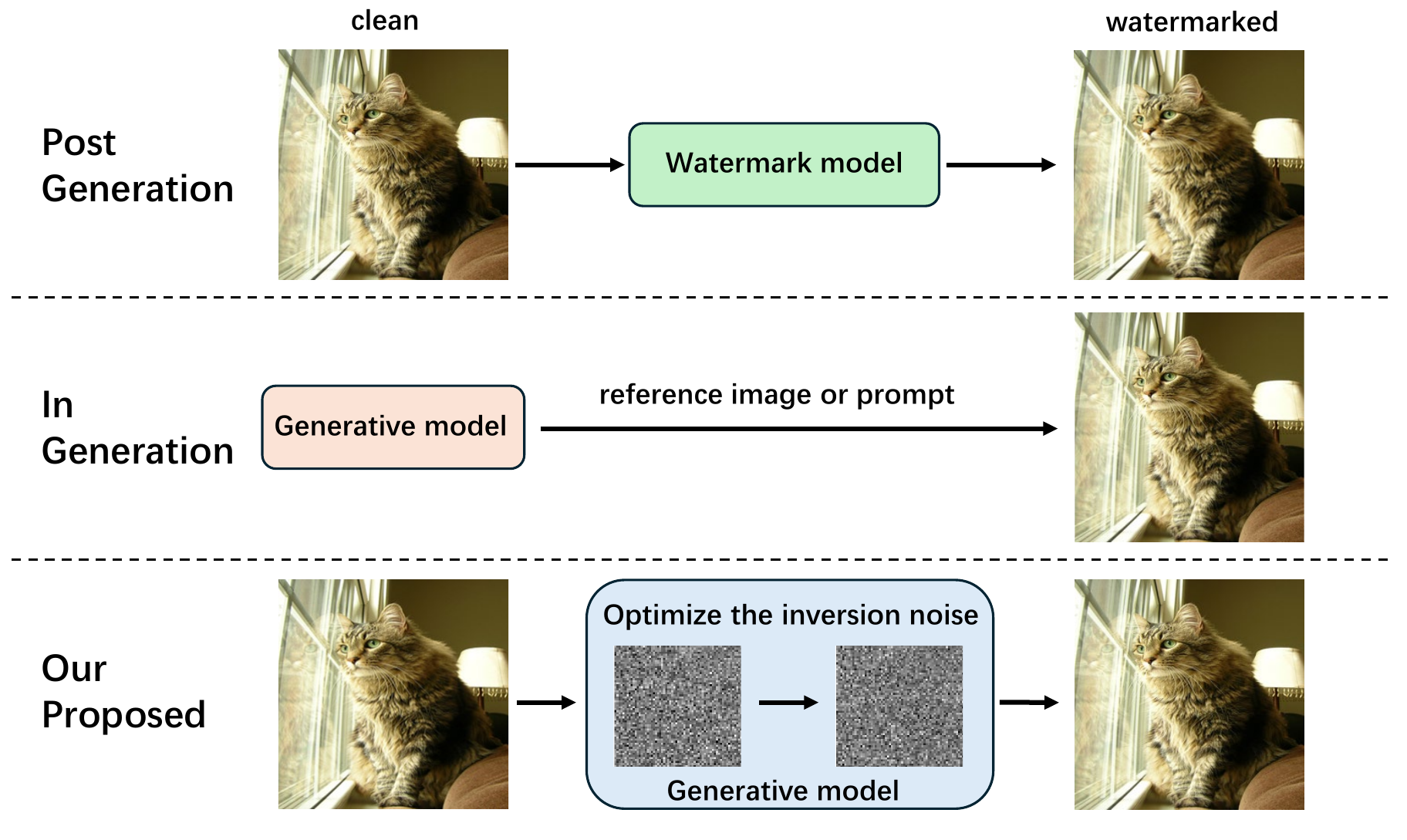}
\caption{Watermarking Algorithm Types. The post-generation method is straightforward to implement but still requires improvements in both security and efficiency, whereas the in-generation method, primarily reliant on generative models, needs further enhancement in watermark imperceptibility. Our method converts a clean image into inversion noise through the null-text inversion process and then embeds the watermark by optimizing this inversion noise.}
\label{motivation}
\end{figure}

Research on existing image watermarking algorithms has made considerable progress and can be broadly categorized into post-generation methods and in-generation methods (as shown in Fig.\ref{motivation}). Post-generation methods embed watermarks into images after they have been generated, typically utilizing techniques such as Discrete Wavelet Transform (DWT) or Singular Value Decomposition (SVD). These algorithms are always straightforward to implement and do not significantly change the original visual quality \citep{pei2022retracted}. However, they may struggle to withstand aggressive modifications or attacks effectively \citep{zhong2020automated}. In contrast, in-generation approaches integrate watermarking directly into the image generation process, often using diffusion models. This method can enhance robustness, as the watermark becomes part of the image's inherent structure \citep{khan2024trade}. However, as the intensity of the watermark increases to improve robustness, a corresponding decrease in image consistency can occur \citep{li2024shallow}. Some methods, such as Tree-Ring \citep{wen2023tree}, may even alter the original semantic information of the image.

Achieving both robustness and imperceptibility is essential for effective watermarking methods, particularly in applications involving copyright protection, digital rights management, and content verification \citep{zhao2024provable}. A robust watermark remains detectable even after the watermarked image undergoes transformations intended to remove or degrade it. For example, if a watermark is easily destroyed by common operations like JPEG compression, it fails to serve its purpose of protecting intellectual property. Imperceptibility is equally essential, especially in media like photographs or videos where visual fidelity is paramount. An overly visible watermark can detract from user experience and undermine its role as a discreet marker of ownership or authenticity \citep{rashid2019designing}. Balancing robustness and imperceptibility is challenging, as enhancing robustness often compromises imperceptibility. For instance, embedding watermarks in more significant bits (MSBs) of pixel values may improve robustness but introduce visible artifacts, while embedding in less significant bits (LSBs) preserves visual quality but offers less resilience against attacks \citep{sharma2024review}. Considering the above challenges, this paper proposes a novel framework that utilizes \textbf{O}ptimizing inversion \textbf{N}oise of diffusion models to obtain high-quality and \textbf{R}obust \textbf{W}atermark (ONRW). The framework diverges from post hoc watermarking methods and other diffusion model-based watermarking approaches. In response to the invisibility requirement of the watermark, we use the null-text inversion process of the diffusion model to map the clean image into inversion noise, and embed the watermark invisible into the image by optimizing the inversion noise method. To prevent the optimization of inversion noise from distorting the original semantics of the image, we specifically introduce self-attention constraints and pseudo-masking strategies. The self-attention constraint guarantees the quality of the watermarked image by maintaining the consistency of the self-attention map, while the pseudo-masking strategy improves the accuracy and robustness of the watermark embedding by only adding the watermark to the foreground objects of the image and filtering out the background areas. \citep{chen2024diffusion}. In response to the robustness requirements of watermarks, our model uses the denoising process of the diffusion model to map the optimized inversion noise into a watermarked image. The iterative denoising process of the diffusion model not only reduces perceptible high-frequency noise, thus enhancing the visual quality of the image but also serves as a powerful purification defense mechanism \citep{nie2022diffusion}, effectively improving the watermark’s robustness against various image watermark attacks. Additionally, we incorporated a simulated attack layer similar to the HiDDeN model \citep{zhu2018hidden} and combined it with the diffusion model’s built-in purification defense mechanism to further enhance the watermark's robustness against image corruptions such as rotation, cropping, image compression, and image reconstruction. 

We evaluate the quality of images generated by our method through both quantitative comparisons and qualitative analyses, and validate its robustness against watermark attacks through extensive experiments, demonstrating the resilience of our encoded watermark information under various forms of image distortions. By measuring the watermark robustness under different image transformations, we show that despite challenges such as size variations, brightness shifts, cropping, JPEG compression, noise attack, and filter attack, our network effectively reconstructs the hidden information encoded in the image. We specifically assess the watermark's resistance to different threat models, including attacks on images compressed by neural autoencoders and watermark attacks based on diffusion models. Experimental results indicate that our watermarking model exhibits strong resistance even against deliberate attacks. 

The main contribution of this paper can be summarized as follows:
\begin{itemize}
    \item We leverage the excellent image generation and purification defense capabilities of the diffusion model's iterative denoising process to propose a novel watermarking framework. This framework optimizes the inversion noise in the latent space of the diffusion model to embed the watermark into the image, allowing for watermark embedding without the need for additional training.
    \item We introduce self-attention constraints, pseudo-mask strategies, and simulated attack layers into the null-text optimization process of the basic diffusion model. This combination not only improves the effectiveness of watermark processing, but also enhances its resistance to various attacks, ensuring that embedded watermarks remain detectable even under challenging conditions.
    \item Extensive experiments demonstrate that our proposed method exhibits higher robustness against various attacks, including image transformations and deliberate distortions, compared to existing watermarking methods.
\end{itemize}

\section{Related Work}

\subsection{Image Generation Models} Image generation models have seen significant advancements over recent years, transforming the landscape of artificial intelligence and creative content generation. The introduction of Generative Adversarial Networks (GANs) by \citet{goodfellow2014generative} represented a seminal development in this domain. GANs generate highly realistic images through the confrontation training of a generator that creates images and a discriminator that evaluates them \citep{saxena2021generative}. Although GANs performed well in generating realistic images, it still faces challenges such as pattern collapse and unstable training. Variational Autoencoders (VAEs) \citep{kingma2014auto} excel in producing diverse outputs but often struggle with the sharpness of generated images compared to GANs. The Vector Quantized-Variational Autoencoder (VQVAE) model has been developed to mitigate these drawbacks, focusing on improving the quality and cohesiveness of the generated imagery \citep{van2017neural}. More recently, the advent of diffusion models has pushed the boundaries of image generation even further.

Early research on diffusion models were grounded in probabilistic models and stochastic processes, such as Langevin dynamics \citep{uhlenbeck1930theory} and diffusion processes. In 2015, \citet{sohl2015deep} proposed the diffusion concept in generative modeling, laying a solid foundation for the subsequent development of diffusion models. \citet{song2019generative,song2020improved} introduced score-based generative models, which have set a new trend in generative modeling due to their remarkable performance. Diffusion models mainly consist of two processes: the forward diffusion and the reverse generation processes. The forward process gradually transforms an image into pure Gaussian noise, while a U-Net architecture is trained to predict the noise added at each timestep in the reverse process. After extensive training on large datasets, diffusion models are capable of generating high-quality images from randomly sampled noise and producing images with specific content based on textual prompts \citep{saharia2022photorealistic,ramesh2022hierarchical,rombach2022high}.

\subsection{Image watermark}
With the rapid development of machine learning and deep learning technologies, watermarking algorithms have made significant progress in copyright protection and content authenticity. \citet{hayes2017generating} proposed an end-to-end learning watermarking scheme, which optimizes the encoder and decoder of the watermark through adversarial training, thereby improving the embedding effect of the watermark. \citet{zeng2023securing} used an encoder to embed the watermark signal into images, enhancing the algorithm's adaptability. \citet{yu2021artificial} proposed a two-stage watermark embedding method, using a trained encoder to generate watermarked images. However, the robustness of these methods needs to be improved, especially in the face of various attack models, the integrity of watermarking is easy to be damaged, thus weakening its reliability as a content authenticity verification tool. In recent years, the emergence of diffusion models has injected new vitality into advancements in image generation technology. \citet{fernandez2023stable} proposed a watermarking method termed Stable Signature that fine-tunes the latent decoder of the stable diffusion model with a binary signature as the condition. \citet{zhao2023recipe} conducted watermark data training in an unconditional diffusion model. In addition, \citet{wen2023tree} proposed the "Tree-Ring Watermark" technique, which embeds tree-ring-like watermarks into the frequency domain of the initial noise in the diffusion model to achieve high concealment and robustness. The image quality generated by these methods needs to be improved, and some watermarks generated during the generation process will also change the original semantic of the image, which is is not suitable for applications that require preserving the original image content.

\subsection{Watermark Evaluation}

The evaluation of digital watermarking primarily focuses on visual quality, detection of false positive rates, and robustness. Early evaluation methods mainly considered the impact of image processing operations on watermarks. For example, \citet{pitas1998method} evaluated the robustness of watermarking methods through ROC curves, considering image processing operations such as translation, rotation, and resizing. \citet{kutter1999fair} introduced objective methods for assessing watermark visual quality and established standards for a fair evaluation of watermarking technologies. These methods primarily focused on the resistance of watermarking technologies to distortions in natural scenes. However, recent research has begun to consider the watermarking resistance to deliberate tampering. \citet{fernandez2023stable} specifically examined the performance of watermarks under two threat models: image-level and generation-model-level. Additionally, \citet{zhao2025invisible} proposed a diffusion model-based watermark attack algorithm to assess the robustness of watermarking models against various attacks. These advancements not only drive progress in the practical application of digital watermarking technology but also offer new perspectives and methods for evaluating and improving watermarking techniques.

\section{Method}\label{sec3}
\subsection{Preliminary}

The vanilla inversion method (such as DDIM inversion) results in deviations that accumulate during the sampling process and are amplified with increasing weight. The final generated noise no longer follows a Gaussian distribution, making it difficult to recover the original image from this noise. To solve this problem, \citet{mokady2023null} proposed the null text inversion method. This method optimizes unconditional text embedding based on DDIM inversion, which can effectively reduce errors in the inversion process. 

The specific steps of null-text inversion are as follows. First, use DDIM inversion to generate a diffusion trajectory from a pure Gaussian noise vector to the image encoding, which serves as an initial reference point for optimization. Then, based on the initial noise trajectory, we adjust the unconditional text embedding via null-text optimization to minimize the inversion error. Specifically, for each time step \(t\), we optimize the unconditional text embedding \(\phi_t\) to minimize the reconstruction error:

\begin{equation}
 \min_{\phi_t}\sum_{t=1}^T\|x_{t-1}-x'_{t-1}(x_t,\phi_t,\Psi(P)\|^2_2 
\end{equation}

where \( x_{t-1} \) is the pivot trajectory obtained from the initial DDIM inversion, \( \phi_t \) is the unconditional embedding, \( \Psi(P) \) is the embedding of the text condition and \( x'_{t-1}\ (x_t, \phi_t, \psi(P)) \) represents the intermediate result obtained through DDIM sampling steps. By optimizing the unconditional embedding, accurate reconstruction of the input image can be achieved without adjusting the model weights or conditional text embedding.

\begin{figure*}[h]
\centering
\includegraphics[width=0.95\textwidth]{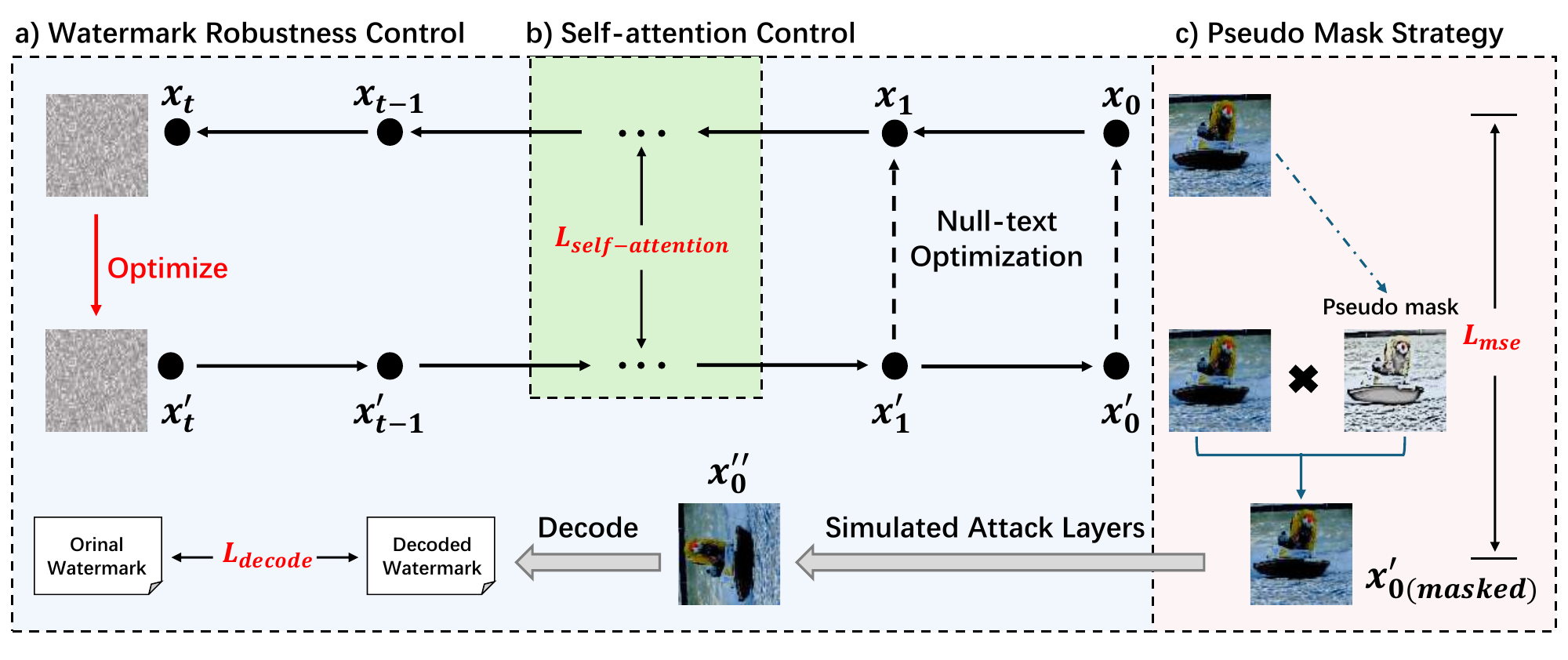}
\caption{Framework of ONRW. We adopt Stable Diffusion and leverage null-text inversion to convert the clean image into inversion noise. The noise is optimized to embed the watermark information. Self-Attention and MSE loss controls ensure concealment, while decoder optimization enhances robustness. Areas are color-coded: blue for Watermark Robustness Control, green for Self-Attention Control, and pink for Pseudo Mask Strategy.}
\label{Framework of ONRW}
\end{figure*}

\subsection{ONRW Framework}
As shown in Fig.\ref{Framework of ONRW}, our proposed ONRW framework is based on the open-source Stable Diffusion model \citep{rombach2022high}. Utilizing the null-text inversion technology \citep{mokady2023null}, we transform the original image $x_0$ to be watermarked into noise:
\begin{equation}
x_t=\underbrace{Inverse \circ \cdots \circ Inverse}_t\left(x_0\right)
\end{equation}
where Inverse(·) denotes the null-text inversion operation. We optimize this inversion noise $x_t$ to $x'_t$ in the latent space to embed the watermark information and then convert it back to the image through a diffusion process. We encourage the watermark information from the watermarked image that has been disrupted by the simulated attack layer.

To achieve a robust and imperceptible watermark embedding framework, we consider the following optimization objectives: (1) Watermark robustness control: minimizing the difference between the input message and the decoded message; (2) Self-attention control: minimizing the difference in self-attention maps before and after optimizing the inversion noise; (3) Pseudo mask strategy: minimizing the pixel difference between the watermarked image and the encoded image.

\subsubsection{Watermark robustness control}

\citet{nie2022diffusion} emphasizes the powerful ability of the inverse process in the diffusion model to combat purification. Utilizing the null-text inversion technology \citep{mokady2023null}, our watermarking framework embeds watermark information into inversion noise and undergoes efficient denoising before decoding into images, which inherently enhances the robustness against attacks such as image transformation, reconstruction, and compression. In addition, inspired by existing work \citep{zhu2018hidden}, we also introduce simulated attack layers including Identity layer, Crop layer, Rotation layer, Resize layer, Brightness layer and JPEG layer. Based on this setting, the watermarked images we generate are input to a randomly applied image distortion attack layer, which helps the model to more robustly handle watermark attacks caused by image transformations in real-world scenarios.

During training, the decoder $D$ extracts watermark information from the watermarked images, which have been generated from the optimized inversion noise through the diffusion denoising process and then processed by the simulated attack layer $W$. The optimization objective is to extract watermark information $k$ embedded in the image as accurately as possible. This process can be formalized as follows:

\begin{equation}
\arg\min_{x_t}L_{decode} = \|D(x''_0)-k\|^2_2,\ where\ x''_0=W(x'_0)
\end{equation}

\subsubsection{Self-attention control}
Given that self-attention maps in diffusion models reflect structural information of images to some extent \citep{chen2024diffusion}, we propose maintaining the consistency of self-attention maps when embedding watermark information into the diffusion process. This approach will help reduce structural differences between the watermarked image and the original image, thereby enhancing visual quality. Specifically, we first create and fix a copy \( x_{t_{(fix)}} \) of the inversion noise and compute the self-attention maps for this copy and the optimized inversion noise \( x_t \), denoted as \( S_{t_{(fix)}} \) and \( S_t \), respectively. To keep \( S_t \) and \( S_{t_{(fix)}} \) as consistent as possible, we introduce self-attention control at all denoising steps, which can be formulated as:
\begin{equation}
\arg\min_{x_t}L_{self-attention} = \|S_t-S_{t(fix)}\|^2_2
\end{equation}
Since \( x_{t_{(fix)}} \) effectively reconstructs the original clean image, this approach allows us to preserve the structural features of the image, making them less affected by the embedded watermark information \citep{song2020denoising}.

\subsubsection{Pseudo mask strategy}
To further optimize the visual quality of watermarked images, we propose embedding watermark information only into the foreground objects of the image, thereby constraining the areas affected by the watermark. Additionally, because various non-human or human image transformations tend to avoid damaging the foreground objects and maintain the semantic content of the image, embedding watermark information solely in the foreground objects also helps enhance the watermark's robustness. Thanks to the ability of cross-attention maps in diffusion models to reveal which parts of the image are closely related to the input guiding text \citep{hertz2022prompt}, we propose generating a "pseudo" mask based on these maps to constrain where the watermark information is added. Specifically, we first aggregate the cross-attention maps between image pixels and C(the caption of the original image) throughout all denoising steps and compute their average:

\begin{equation}
P={Average}\left({Cross}\left(x_t, t, C\right)\right)
\end{equation}

\begin{equation}
M_{{soft }}={Up}\left(\frac{P}{{Max}(P)}\right)
\end{equation}

\begin{equation}
{(Optional)}M_{{hard}}=\left\{  
             \begin{array}{lr}  
             1, &  M_{{soft }}>0.5 \\
             
             0, & M_{{soft }} \leq 0.5\\      
             \end{array}  
	\right. 
\label{hard-mask}
\end{equation}
Here, Cross(·) represents the cumulative cross-attention maps during denoising with the Stable Diffusion model. Up(·) resizes the cross-attention map through upsampling due to the existing encoder downsampling in the Autoencoder and U-Net. Max(·) extracts the maximum value and normalizes the cross-attention maps P.  Given that the normalized \(M_{soft}\) ranges from 0 to 1, Eq. (\ref{hard-mask}) can be utilized optionally to derive a hard mask. Therefore, this "pseudo" mask filters out background areas, adding the watermark only to foreground objects, as formulated:

\begin{equation}
x'_{0(masked)} = x'_0 \times M + x_0 \times (1-M)
\end{equation}
where $x'_0=\underbrace{Denoise \circ \cdots \circ Denoise}_t\left(x_t\right)$ and Denoise(·) represents the diffusion denoising process under the null-text inversion method. This approach effectively hides the watermark information to ensure its invisibility. To measure the difference between the original image and the watermarked image, we use the mean squared error (MSE) loss function:
\begin{equation}
\arg\min_{x_t}L_{mse} = \|x'_{0(masked)}-x_0\|_2
\end{equation}

The final objective function of our method is as follows, where \( \alpha \), \( \beta \), and \( \gamma \) represent the weight factors for each loss term:

\begin{equation}
\arg\min_{x_t}L = \alpha L_{decoded}+\beta L_{self-attention}+\gamma L_{mse}
\label{total_loss}
\end{equation}

\section{Experiments}\label{sec4}
\subsection{Experimental Setting}

We use DDIM \citep{song2020denoising} as the sampler of Stable Diffusion-v2 \citep{rombach2022high} and choose the decoder of HiDDeN \citep{zhu2018hidden} as our watermark extractor. Our approach does not require additional training, which eliminates any training costs. Both the DDIM sampler and HiDDeN decoder are sufficiently pretrained to effectively extract watermarks without the need for retraining. During the inference phase, considering the time resources required for the null-text inversion process, we performed 120 inference steps on a single NVIDIA GeForce RTX 4090 to generate and detect the watermark, which took approximately one minute per image (see Fig.\ref{ablation}). In order to achieve reversibility of image transformation, we set the guidance ratio of stable diffusion to 4.5. We use the AdamW optimizer \citep{loshchilov2017deCoupled} to optimize the inversion noise \(x_t\), with the learning rate set to \(1 \times 10^{-2}\). The weight ratio \(\alpha\), \(\beta\) and \(\gamma\) (\ref{total_loss}) in the loss function equation is set to 100:80:20. All experiments were performed on a single NVIDIA GeForce RTX 4090.

To maintain consistency in comparative experiments, we selected the COCO2014 dataset \citep{lin2014microsoft} and ImageNet2012 dataset \citep{deng2009imagenet} as the basis, with all images resized to 256x256 pixels, aiming to evaluate the quality of generated images and watermark embedding effects. For watermark embedding, we chose a watermark information length of 48 bits for each experiment, with each bit randomly and uniformly sampled. 

\subsection{Image Quality}

\subsubsection{Qualitative Comparison}

\begin{figure}[h]
\centering
\includegraphics[width=1.0\columnwidth]{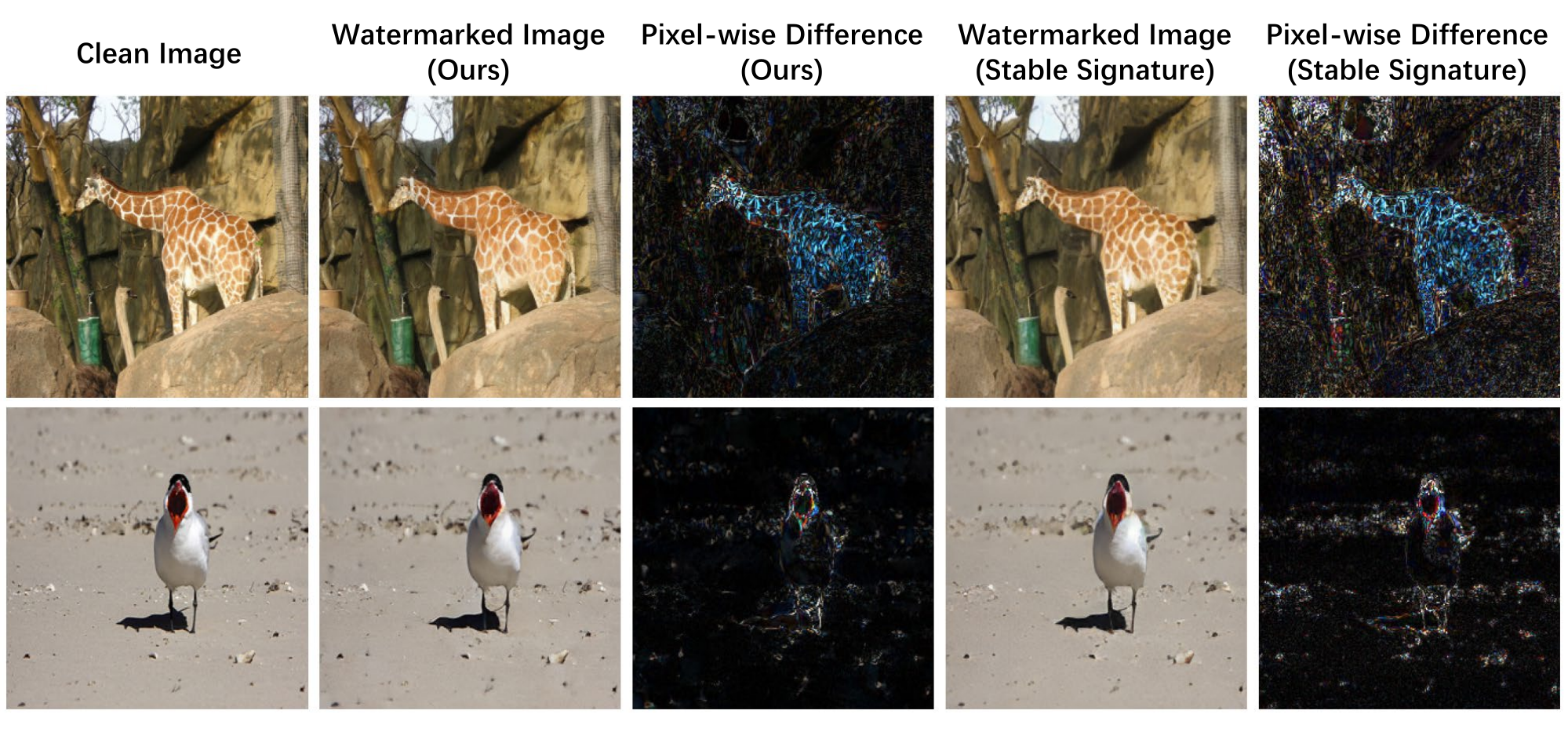}
\caption{Watermarked image and pixel-level differences from the original image (magnified three times for enhanced visual clarity) on COCO dataset. Our method embeds the watermark in key areas with minimal impact on the background.}
\label{Pixel Difference Comparison}
\end{figure}

Considering the complexity and subjective experience of the human visual system, we first conducted a qualitative analysis of the images generated by the watermarking method. To observe which spatial locations the watermark information has been embedded in, we present the pixel-level differences between the watermarked image and the original image in Fig.\ref{Pixel Difference Comparison}. For visual clarity, we magnified the pixel difference maps by three times. From the magnified pixel difference map, it is clear that our method, compared to the Stable Signature approach, successfully embeds the watermark into the foreground objects of the image, thereby preserving the uniformity and consistency of the background region. The characteristic of embedding the watermark into the foreground objects of the image has two main advantages: it minimizes the impact on the overall appearance, thus maintaining the aesthetic quality of the image, and it enhances robustness against corruption in the background regions. 

\begin{figure}[h]
\centering
\includegraphics[width=1.0\columnwidth]{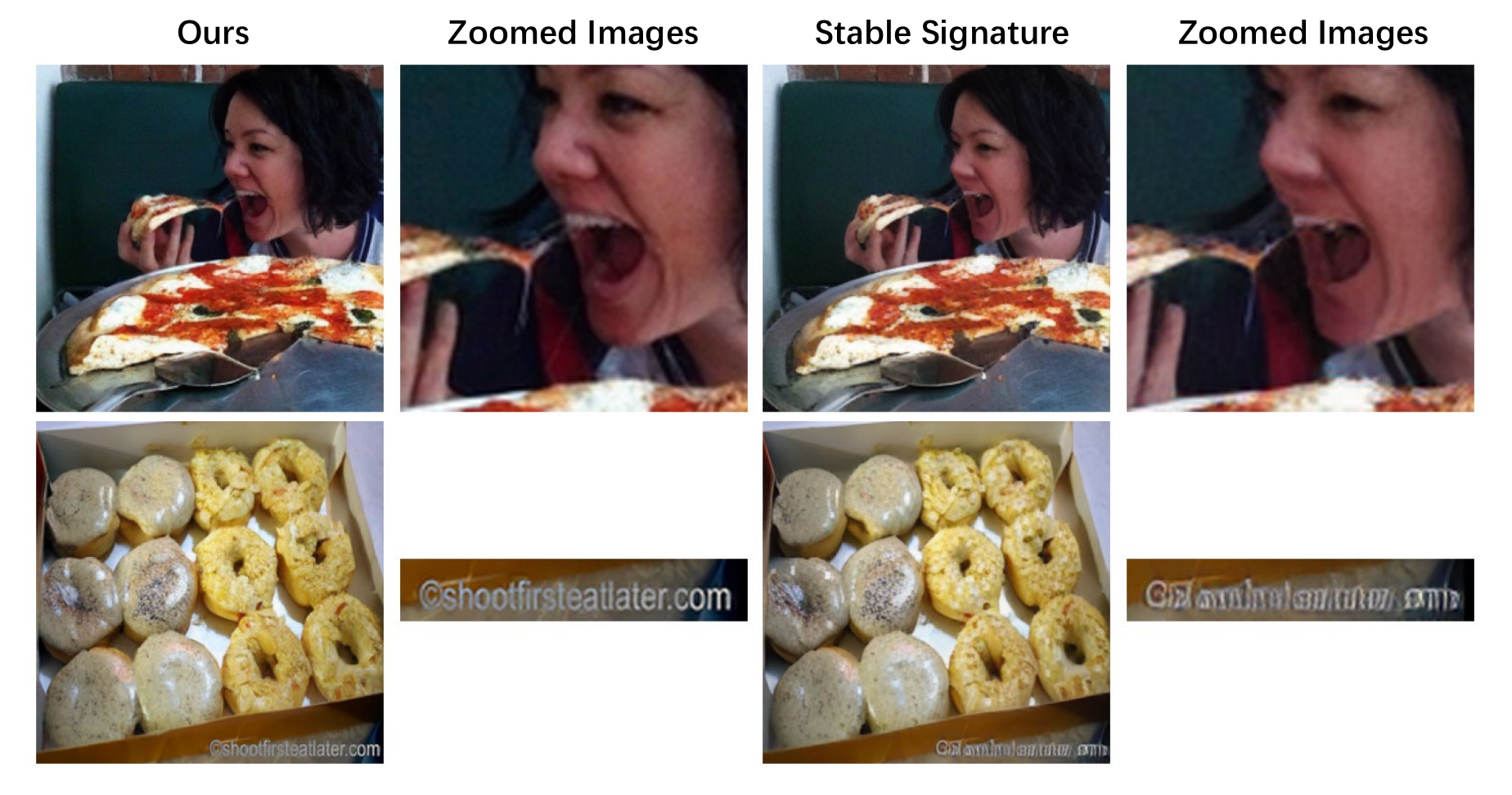}
\caption{Qualitative comparison of watermarked images generated by our method and the existing generative model-based watermarking method Stable Signature.}
\label{word-face}
\end{figure}

To visually observe the impact of watermark information on image quality, we present a schematic of the magnified watermarked image in Fig.\ref{word-face}. Watermarked images generated by Stable Signature are more prone to issues such as facial feature blurring and text distortion, which directly affect the visual quality of the images. 

We also qualitatively compared the quality of watermarked images generated by our model with other watermarking models in Appendix C, including Dct-Dwt \citep{cox2007digital}, Dct-Dwt-Svd \citep{cox2007digital}, RivaGAN \citep{zhang2019robust}, SSL Watermark \citep{fernandez2022watermarking}, HiDDeN \citep{zhu2018hidden}, Stable Signature \citep{fernandez2023stable}, and Tree-Ring \citep{wen2023tree}. The comparative results show that our method embeds robust watermarks into images while preserving the image information. 

\subsubsection{Quantitative Evaluation}

To ensure the objectivity and accuracy of the comparison results, we also quantitatively evaluate the invisibility of the watermarks generated by our method. For a fair comparison, we selected two latest methods that also use diffusion models for watermark generation as benchmarks: the Stable Signature model \citep{fernandez2023stable} and the Tree-Ring model \citep{wen2023tree}. Tab. \ref{table1} presents the quantitative evaluation results.

\begin{table}[h]

\centering
\resizebox{1.0\columnwidth}{!}{%
\begin{tabular}{@{}clccccc@{}}
\hline \hline
Dataset & Model & PSNR\footnotemark[1] & SSIM\footnotemark[1] & Linf & LPIPS\footnotemark[1] & MSE\footnotemark[1] \\
\midrule
\multirow{4}{*}{COCO} & Stable Signature & 25.55 & 0.75 & 155.78 & 0.0590 & 233.48 \\
 & Tree-Ring(text-to-image) & 7.53 & 0.20 & 254.59 & 0.606 & 12235.57 \\
 & Tree-Ring(image-to-image) & 10.07 & 0.25 & 246.82 & 0.656 & 6983.25 \\
 & Ours & \textbf{27.11} & \textbf{0.88} & \textbf{140.59} & \textbf{0.0569} & \textbf{148.66} \\
\midrule
\multirow{4}{*}{ImageNet} & Stable Signature & 26.13 & 0.76 & 146.72 & 0.0596 & 217.48 \\
 & Tree-Ring(text-to-image) & 9.21 & 0.16 & 246.36 & 0.661 & 8563.07 \\
 & Tree-Ring(image-to-image) & 9.64 & 0.24 & 241.75 & 0.676 & 8011.00 \\
 & Ours & \textbf{27.83} & \textbf{0.89} & \textbf{130.76} & \textbf{0.559} & \textbf{132.40} \\ 
\hline \hline
\end{tabular}}
\caption{Generation quality comparison of the diffusion-based watermarking models. }
\label{table1}%
\end{table}

The evaluation results show that our method excels in all four metrics on both COCO dataset and ImageNet dataset. Specifically, our method outperforms the Stable Signature and Tree-Ring methods with PSNR score improvements of 1.56 and 17.04 on COCO dataset, respectively. However, overall, the PSNR values for our method, as well as the Stable Signature and Tree-Ring methods, are relatively low. This may be caused by the inherent distribution changes introduced when the existing diffusion model first adds noise and then denoises to reconstruct the image. Even though visually, there is almost no difference between the image reconstructed using null-text inversion and the original image, commonly used image quality evaluation metrics (such as PSNR and SSIM) are often sensitive to such distribution changes. 

\begin{table}[h]
\centering
\renewcommand{\arraystretch}{1.15}
\resizebox{1.0\columnwidth}{!}{%
\begin{tabular}{@{}llllll@{}}
\hline \hline
                                       & PSNR  & SSIM & Linf   & LPIPS & MSE \\
\hline
Null-text Inversion                    & 26.48 & 0.76 & 188.81 & 0.0539 & 163.78 \\
Null-text Inversion + Watermark (Ours) & 27.11 & 0.88 & 140.59 & 0.0569 & 148.66 \\
\hline \hline
\end{tabular}}
\caption{Comparison of PSNR Values Between Null-text Inversion Reconstructed Images and Watermarked Images Against the Original Images.}
\label{null-text}%
\end{table}

As shown in Tab. \ref{null-text}, the PSNR value for the image obtained using only null-text inversion without watermarking is 26.48. On this basis, our method uses attention control, pseudo mask, and other strategies to embed a watermark, resulting in a PSNR value of 27.11. Therefore, the image inversion method may be the main factor restricting the image quality metrics in our framework. How to design a better image inversion method may be a future research direction, but it is not the focus of this paper, nor do we consider it a major limitation. This is because the visual difference between the watermarked image and the original image is negligible, which meets the need for watermark concealment.

\subsection{Robustness against Image Transformations}
This subsection evaluates the robustness of the generated watermarks under various image transformations. Specifically, the transformations tested included cropping (ratios of 0.1, 0.5), rotations (25 degrees, 90 degrees), size changes (zooming to 0.3x, 0.7x), brightness adjustments (1.5x, 2x), JPEG Compression (compression quality 80, 50), Uniform noise (attack with values ranging from -10 to 10) and Mean filter attack (kernel size of 3). These transformations encompass typical geometric and photometric edits, effectively evaluating the watermark's performance under various practical scenarios. Through these tests, we can understand the visibility and durability of watermarks under extreme image transformation conditions. Considering the significant changes to the original image generated by the Tree-Ring watermark and the dependency of these changes on the prompt content, we excluded this model in subsequent robustness tests.

We use bit accuracy as the primary metric to evaluate the robustness of watermarks to evaluate the accuracy of watermark extraction. Higher bit accuracy indicates better recovery performance of the watermark under various distortion conditions, and thus greater robustness. Our comparison objects include post-generation methods (Dwt-Dct, Dwt-Dct-Svd, SSL Watermark, HiDDeN, RivaGAN) and the in-generation method (Stable Signature).

\begin{table*}[h]
\centering
\resizebox{\textwidth}{!}{
\begin{tabular}{@{}clccccccc@{}}
\hline \hline
Dataset & Attack & Dwt-Dct & Dwt-Dct-Svd & SSL Watermark & HiDDeN & RivaGAN & Stable Signature & Ours \\ \midrule
\multirow{14}{*}{COCO} & None & 0.93 & 0.99 & 0.99 & 0.99 & 0.84 & 0.99 & \textbf{0.99} \\
 & Crop\_01 & 0.49 & 0.50 & 0.53 & 0.88 & 0.61 & 0.92 & \textbf{0.99} \\
 & Crop\_05 & 0.53 & 0.52 & 0.83 & 0.97 & 0.76 & 0.99 & \textbf{0.99} \\
 & Rot\_25 & 0.47 & 0.51 & 0.91 & 0.61 & 0.61 & 0.64 & \textbf{0.93} \\
 & Rot\_90 & 0.63 & 0.53 & \textbf{0.97} & 0.58 & 0.52 & 0.51 & 0.94 \\
 & Resize\_0.3 & 0.48 & \textbf{0.99} & \textbf{0.99} & 0.54 & 0.58 & 0.65 & 0.96 \\
 & Resize\_0.7 & 0.72 & 0.99 & 0.99 & 0.87 & 0.79 & 0.96 & \textbf{0.99} \\
 & Brightness\_1.5 & 0.39 & 0.46 & 0.63 & 0.99 & 0.53 & 0.98 & \textbf{0.99} \\
 & Brightness\_2.0 & 0.64 & 0.50 & 0.56 & 0.98 & 0.45 & 0.96 & \textbf{0.98} \\
 & JPEG\_80 & 0.44 & 0.52 & 0.73 & 0.93 & 0.56 & 0.90 & \textbf{0.97} \\
 & JPEG\_50 & 0.42 & 0.51 & 0.59 & 0.83 & 0.51 & 0.84 & \textbf{0.89} \\
 & Noise & 0.49 & 0.53 & \textbf{0.99} & \textbf{0.99} & 0.60 & \textbf{0.99} & 0.98\\
 & Filter & 0.53 & 0.98 & \textbf{0.99} & 0.69 & 0.81 & 0.90 & 0.91 \\
 & Average & 0.55 & 0.66 & 0.82 & 0.83 & 0.63 & 0.86 & \textbf{0.96}\\ \midrule
\multirow{14}{*}{ImageNet} & None & 0.86 & 0.99 & 0.99 & 0.99 & 0.78 & 0.99 & \textbf{0.99} \\
 & Crop\_01 & 0.51 & 0.50 & 0.59 & 0.87 & 0.61 & 0.91 & \textbf{0.98} \\
 & Crop\_05 & 0.51 & 0.53 & 0.80 & 0.97 & 0.72 & 0.98 &\textbf{0.99} \\
 & Rot\_25 & 0.51 & 0.51 & 0.85 & 0.61 & 0.58 & 0.64 & \textbf{0.90} \\
 & Rot\_90 & 0.49 & 0.54 & \textbf{0.92} & 0.58 & 0.52 & 0.50 & 0.91 \\
 & Resize\_0.3 & 0.57 & 0.98 & 0.66 & 0.54 & 0.58 & 0.63 & \textbf{0.98} \\
 & Resize\_0.7 & 0.73 & 0.99 & 0.87 & 0.86 & 0.73 & 0.95 & \textbf{0.99} \\
 & Brightness\_1.5 & 0.41 & 0.45 & 0.92 & 0.98 & 0.53 & 0.97 & \textbf{0.98} \\
 & Brightness\_2.0 & 0.49 & 0.50 & 0.83 & \textbf{0.97} & 0.49 & 0.95 & 0.96 \\
 & JPEG\_80 & 0.50 & 0.52 & 0.85 & 0.91 & 0.56 & 0.88 & \textbf{0.96} \\
 & JPEG\_50 & 0.51 & 0.51 & 0.66 & 0.80 & 0.53 & 0.82 & \textbf{0.88} \\
 & Noise & 0.47 & 0.53 & 0.91 & 0.97 & 0.60 & \textbf{0.98} & 0.96 \\
 & Filter & 0.60 & 0.96 & \textbf{0.98} & 0.69 & 0.74 & 0.88 & 0.91 \\
 & Average & 0.55 & 0.65 & 0.83 & 0.83 & 0.61 & 0.85 & \textbf{0.95} \\
\hline \hline
\end{tabular}
}
\caption{Quantitative evaluation of the performance of watermarking models under various image transformations. }
\label{table2}%
\end{table*}

As shown in Tab. \ref{table2}, our method exhibits high robustness across various image transformations on both COCO dataset and ImageNet dataset. Taking the results on the COCO dataset as an example, for common geometric(Crop, Rot, Resize, JPEG) and photometric(Brightness) edits, the average bit accuracy of our method exceeds 0.95. Compared to the recently proposed in-generation method, the Stable Signature method, our approach, which is also based on diffusion models, achieves a 43\% performance advantage when rotated 90 degrees and a 31\% advantage when resized to 0.3x. In contrast to the post-generation method SSL, our approach achieves a 42\% performance advantage for a brightness change of 2x and a 46\% advantage for a crop ratio of 0.1.

\subsection{Robustness against Intentional Manipulations}

Following the setup in existing works \citep{fernandez2023stable}, we both use neural autoencoders for image compression and diffusion models for image reconstructing to evaluate the robustness of the generated watermarks against intentional manipulations.

\subsubsection{Compression of Images Using Neural Autoencoders}

\begin{figure}[h]
\centering
\includegraphics[width=1.0\columnwidth]{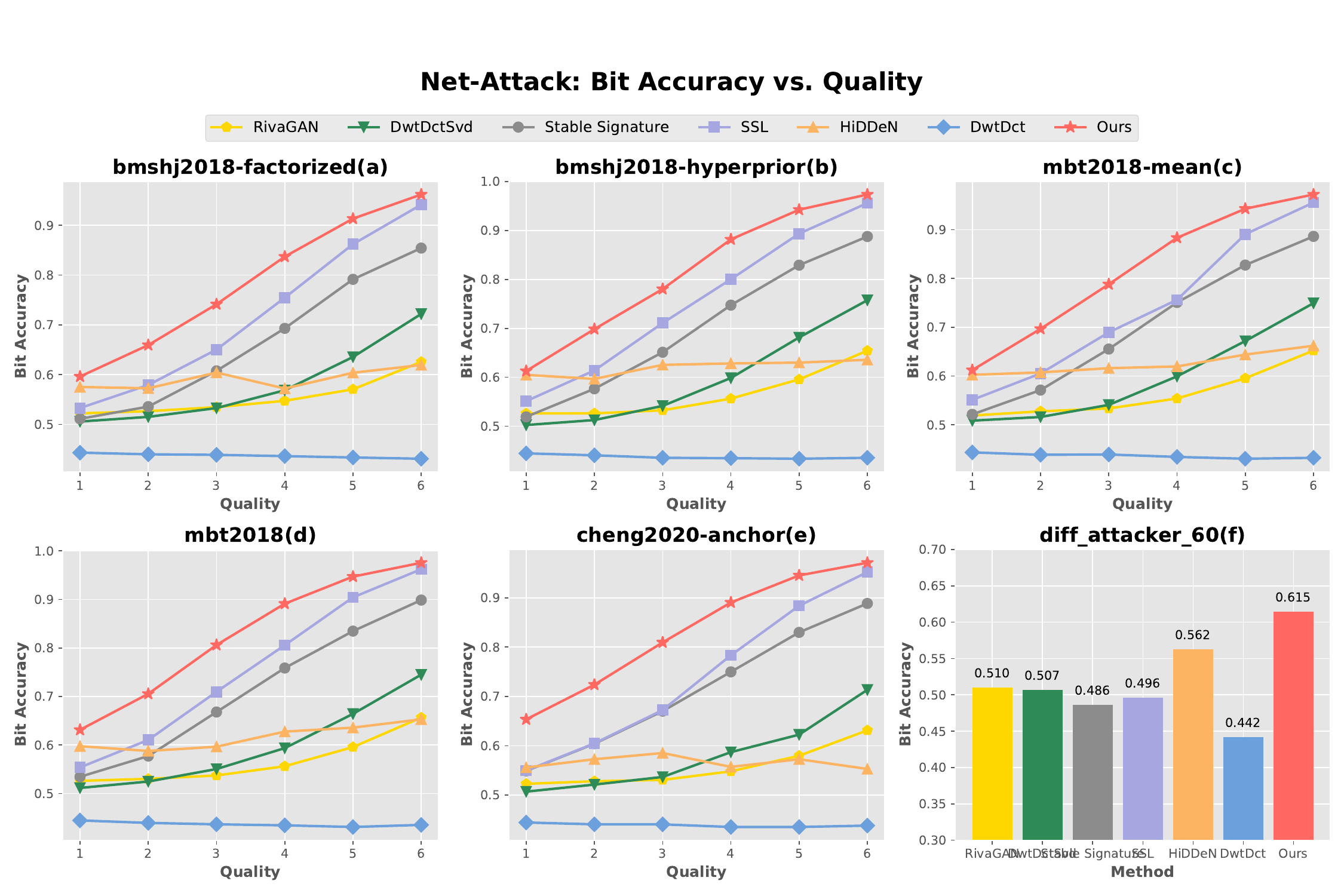}
\caption{Visual comparison of the robustness of various watermark generation models against different types of neural autoencoder attacks under varying compression rates.}
\label{net-attack}
\end{figure}

Neural autoencoders are commonly used deep learning models for image compression and decompression. In watermarked image attacks, autoencoders can be used to attempt to remove or destroy the watermark in an image, making it difficult to detect or extract it. We selected five state-of-the-art autoencoder models from the CompressAI library, including bmshj2018-factorized, bmshj2018-hyperprior \citep{balle2018variational}, mbt2018, mbt2018-mean \citep{minnen2018joint}, cheng2020-anchor \citep{cheng2020learned} and evaluated the robustness of various watermark generation models against different types of neural autoencoder attacks at different compression rates on COCO dataset. The comparison results are shown in Fig.\ref{net-attack}. Generally speaking, for the same encoder, the larger the quality parameter, the higher the compression quality, the more details the compressed image retains, and the smaller the distortion, but the file size of the image will also increase accordingly.

Our method consistency exhibits the highest watermark extraction accuracy across seven different neural autoencoder models. Classic post-generation methods such as HiDDeN, Dwt-Dct and RivaGAN almost lose their robustness when facing compression attacks; even at a lower level of image compression (set to 6), these methods achieve extraction accuracy below 70\%. Although our method, Dwt-Dct-Svd method, SSL Watermark and Stable Signature all show a significant decrease in extraction accuracy with increasing compression rates, out method consistently demonstrates superior robustness compared to others across different compression strengths. When using the compression attack method proposed by \citet{cheng2020learned}, our method achieves a watermark extraction accuracy of 61\% at a compression quality of 4, which is 11\% higher than SSL, 14\% higher than Stable Signature, and 34\% higher than HiDDeN.

\subsubsection{Watermark Attacks Based on Diffusion Models}

Reconstruction attacks based on diffusion models \citep{zhao2025invisible}, which first add random noise to an image to destroy the watermark and then reconstruct the image, have been proven to be an effective watermark attack method. As show in Fig.\ref{net-attack} (f),  reconstruction attacks based on diffusion models indeed cause significant performance degradation across various watermarking algorithms. However, our method outperforms Stable Signature by 12\%, SSL by 11\%, and HiDDeN by 5\%, demonstrating the robustness of our approach.

\subsection{Ablation discussion}

\subsubsection{The Effect of Watermark Robustness Control}

We conducted ablation studies to demonstrate the impact of the simulated attack layer strategy and the null-text optimization strategy. Specifically, we removed the simulated attack layer and the null-text optimization process from our method, respectively, while keeping all other parameters consistent. We then analyzed the quality of the images and the robustness of the watermarks generated by our method on COCO dataset.

\begin{table}[h]
\resizebox{1.0\columnwidth}{!}{%
\begin{tabular}{@{}lcc@{}}
\hline \hline
Model Type & PSNR    & Accuracy on Bit \\ 
\midrule
Orig                & 27.11 & 0.96545         \\
Without Simulated Attack Layer           & 26.06 & 0.86818         \\
Without Null-text Optimization           & 26.58 & 0.97182         \\
\hline \hline
\end{tabular}}
\caption{Ablation of Watermark Robustness Control.}
\label{table4}%
\end{table}

As show in Tab.\ref{table4}, when the simulation attack layer is removed, there is a noticeable decline in both the image quality and the watermark robustness, indicating that this strategy is instrumental in enhancing our method’s performance against various watermark attacks and in improving watermark quality. This may be because the simulated attack layer can adaptively adjust the embed strength and position of the watermark based on the image content and the expected attack type. Such adaptive embedding ensures that the watermarks can resist various attacks as much as possible without destroying the visual quality of the image. Additionally, the removal of the Null-text optimization process results in a decrease in image quality, while the watermark robustness increases, suggesting that the Null-text optimization improves image quality at the expense of watermark robustness.

\subsubsection{The Effect of Pseudo Masks}
We evaluated the impact of different mask types and upsampling strategies on image generation quality (measured by PSNR) and watermark extraction accuracy (measured by bit accuracy). Specifically, the evaluation covered three scenarios: no pseudo mask, soft pseudo mask, and hard pseudo mask. The hard pseudo mask was hardened to convert the pseudo mask into a binary mask. The upsampling strategies included nearest neighbor upsampling and bilinear upsampling. 

\begin{table}[h]
\centering
\renewcommand{\arraystretch}{1.15}
\resizebox{1.0\columnwidth}{!}{%
\begin{tabular}{@{}llll@{}}
\hline \hline
Mask Types            & Upsampling Strategy & PSNR    & Accuracy on Bit \\ 
\midrule
None                  & None                & 22.3386 & 0.99958         \\
{Hard} & Nearest             & 26.6876 & 0.93166         \\
                      & Bilinear            & 26.2167 & 0.93958         \\
{Soft} & Nearest             & 27.3131 & 0.99166         \\
                      & Bilinear            & 27.0484 & 0.99333         \\ 
\hline \hline
\end{tabular}}
\caption{Ablation on Pseudo Masks.}
\label{table3}%
\end{table}

As shown in Tab. \ref{table3}, soft masks outperform hard masks and no masks in terms of image generation quality and bit accuracy. This suggests that during the image generation process, soft masks can produce higher-quality images and provide more accurate bit information. Regarding upsampling strategies, bilinear upsampling and nearest neighbor upsampling perform similarly when using soft masks, with bilinear upsampling slightly enhancing bit accuracy. This indicates that bilinear upsampling might be more suitable for processing soft masks. Consequently, we employed the combination of soft masks and bilinear upsampling in the main experiments.

\subsubsection{The Effect of \(\alpha\), \(\beta\), \(\gamma\) and iteration}

\begin{figure}[h]
\centering
\includegraphics[width=1.0\columnwidth]{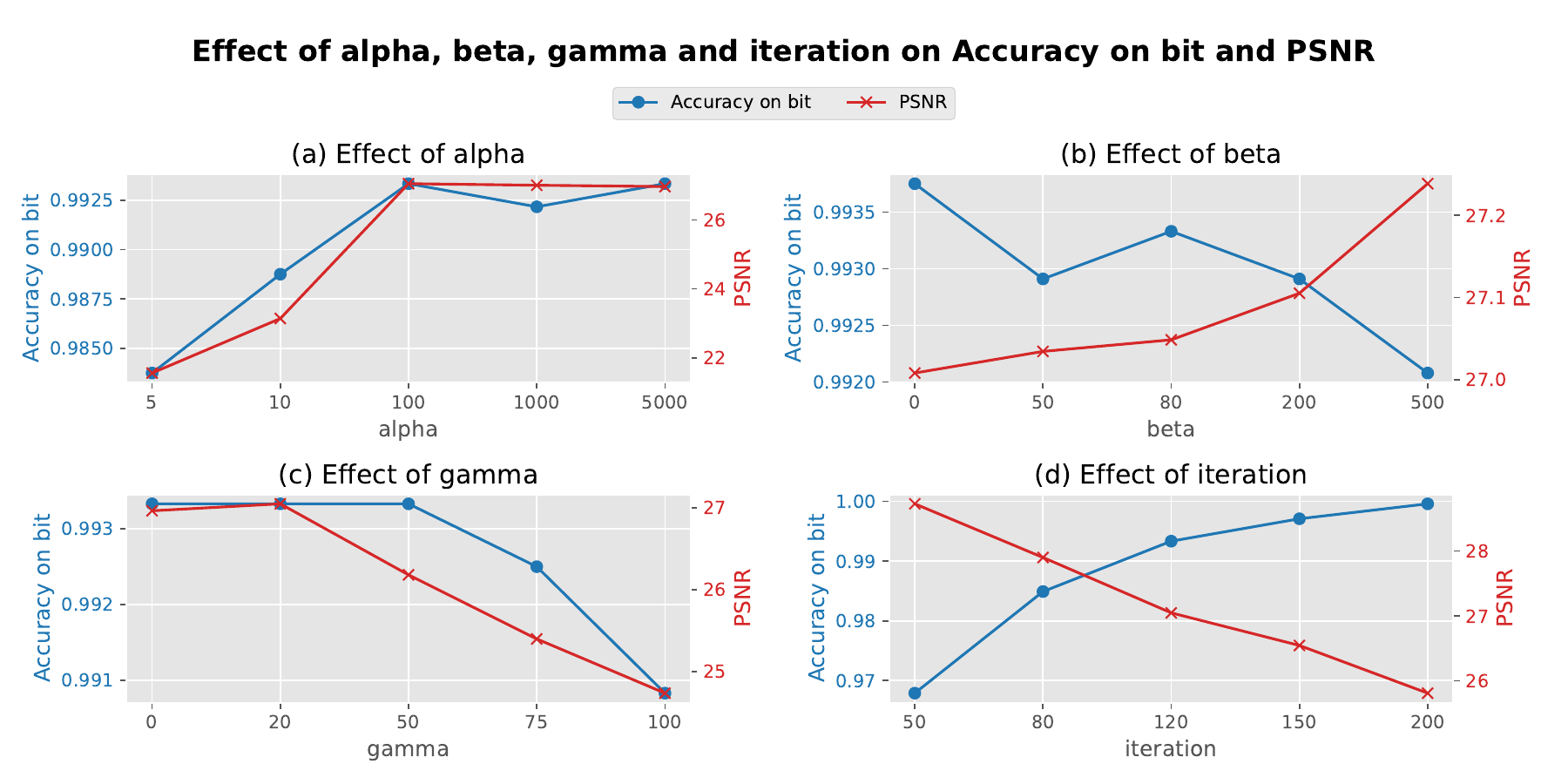}
\caption{Ablation on $\alpha$, $\beta$, $\gamma$ and iteration.}
\label{ablation}
\end{figure}

In this subsection, we investigated the impact of hyperparameters \(\alpha\), \(\beta\), \(\gamma\), and the number of iterations on bit accuracy and peak signal-to-noise ratio (PSNR). Fig.\ref{ablation} illustrates how variations in each hyperparameter impact the performance metrics.

As illustrated in Fig.\ref{ablation}, an increase in the \(\alpha\) value results in a general upward trend in watermark extraction accuracy and peak signal-to-noise ratio (PSNR). However, when $\alpha$ becomes sufficiently large, PSNR and bit accuracy remain almost unchanged. This indicates that moderately increasing $\alpha$ can enhance the robustness and accuracy of the model, but excessively high $\alpha$ values may not yield further significant improvements. In contrast, the \(\beta\) value generally exhibits a negative correlation with watermark extraction accuracy and a positive correlation with PSNR. Changes in the \(\gamma\) value significantly affect both bit accuracy and PSNR. When \(\gamma\) increases from 0 to 20, PSNR increases while maintaining high bit accuracy. When \(\gamma\) exceeds 50, both bit accuracy and PSNR decrease significantly. This suggests that a moderate increase in \(\gamma\) can improve model performance, but an excessively high \(\gamma\) value can cause performance degradation. Based on these findings, we selected \(\alpha\), \(\beta\), and \(\gamma\) values of 100, 80, and 20, respectively.

Additionally, Fig.\ref{ablation}(d) shows that increasing the number of iterations enhances watermark extraction accuracy but compromises image quality. Since more iterations require longer optimization times, we balanced time consumption, image quality, and attack robustness by setting the number of iterations to 120.

\section{Conclusion}

In this work, we propose a novel method for embedding watermarks into images based on an inversion process using diffusion models. Compared to existing works, the watermark embedding approach proposed in this paper fully leverages the diffusion model's iterative denoising process to mitigate high-frequency perceptible noise and purify the embedded information, thereby enhancing both the visual quality and robustness of the watermark. The experimental results demonstrate that our watermarking method exhibit superior robustness compared to existing approaches under both image corruption scenarios and malicious attacks. We hope our work paves the way for further research on watermark generation based on diffusion models.

\bibliography{aaai25}

\end{document}

\maketitle
\appendix

\section{Model Framework Details} 
\subsection{Watermark encoder/extractor}
We selected stable diffusion v2 as the base model. In the framework, the watermark extractor utilizes the pre-trained and whitened HiDDeN model following the settings outlined in stable signature \citep{fernandez2023stable}.

\subsection{Simulated attack layer}
The simulated attack layer in our model includes the following types of image transformations:

\begin{itemize}
    \item Identity Layer: This layer passes the image through unchanged to simulate no transformation.
    \item Crop Layer: This layer creates random square crops of the image, extracting sections at varying positions and sizes to simulate partial occlusion or incomplete watermarked images.
    \item Rot Layer: This layer rotates the image by a specified angle to simulate angular variations, enhancing robustness against rotational transformations.
    \item Resize Layer: This layer simulates changes in the image size, improving the model’s adaptability to different resolutions.
    \item Brightness Layer: This layer modifies the image brightness to train the watermarking model for resilience against varying lighting conditions.
    \item JPEG Layer: This layer simulates compression artifacts from image transmission or storage. Since JPEG compression is non-differentiable but crucial, we employ the JPEG compression layer proposed by MBRS \citep{jia2021mbrs} for this simulation.
\end{itemize}

Each attack simulation layer has a scalar hyperparameter to control the distortion intensity, such as the pixel retention ratio \( p \) for the Crop layer and the compression quality factor \( q \) for the JPEG layer. 

\section{Implementation Details}
\subsection{Watermarking methods}
The watermarking models involved in this article include post-generation methods (Dwt-Dct, Dwt-Dct-Svd, SSL Watermark, HiDDeN, RivaGAN) and the in-generation method (Stable Signature, Tree-Ring). For Dct-Dwt and Dct-Dwt-Svd, we used the implementation available at https://github.com/ShieldMnt/invisible-watermark (the implementation used in stable diffusion).  For HiDDeN, we use pre-trained and whitened HiDDeN models from stable signature paper (to ensure consistent dimensions of processed watermarked images). For SSL watermark, RivaGAN and stable signatures, we use the default pretrained model from the original paper. For Tree-Ring, the officially provided model is mainly used to generate images (Vensen diagrams) from text. To ensure a consistent and fair comparison, we also reproduced an image-to-image version of Tree-Ring. For simplicity, we added watermarks to the images one by one in all cases. In practice, watermarking can be performed in batches, which is more efficient.

\subsection{Image transformations}
\begin{figure}[h]
\centering
\includegraphics[width=1.0\columnwidth]{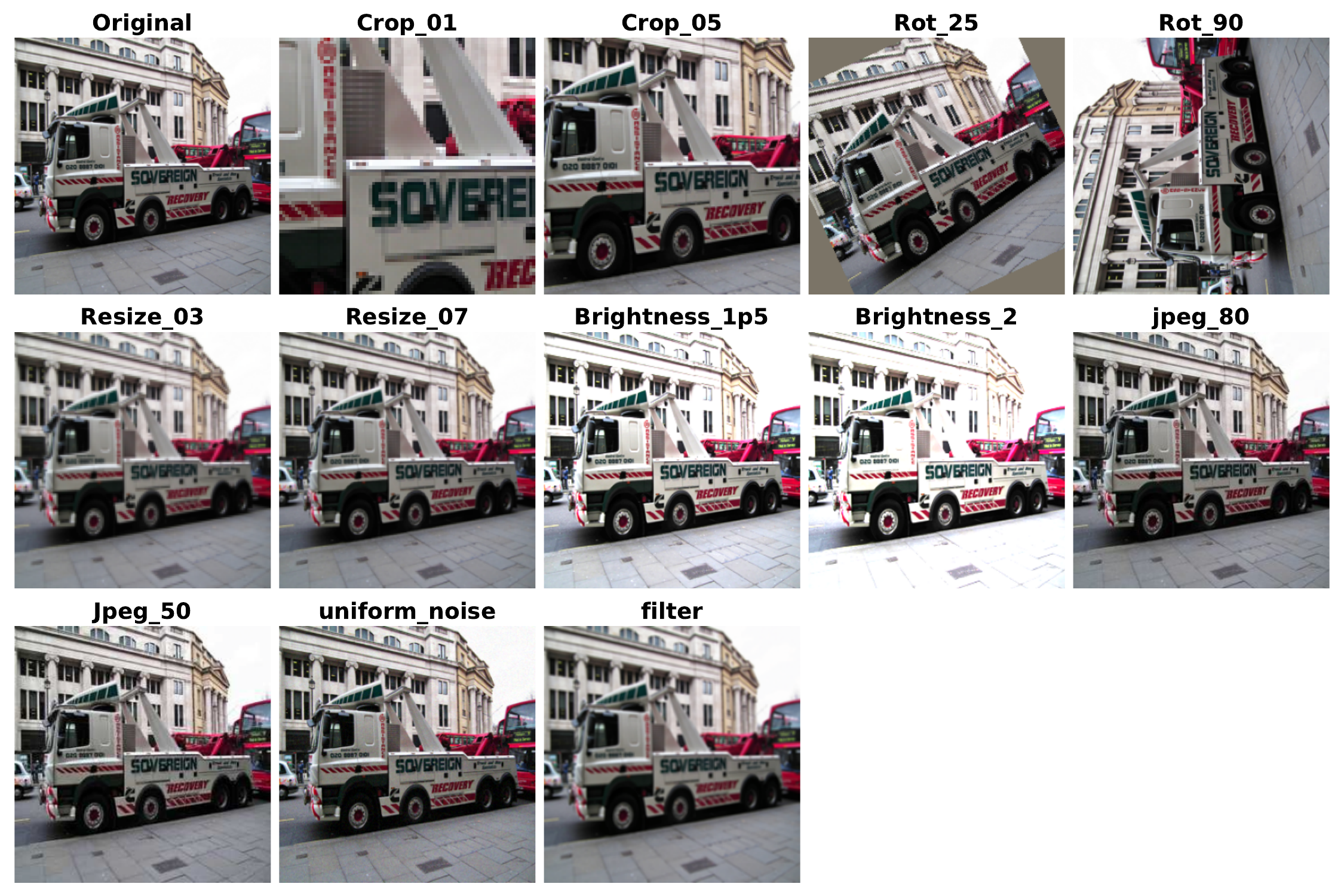}
\caption{The visual effects of the same image under different image transformations.}
\label{transformation}
\end{figure}

We evaluated the robustness of the watermark against image transformations, primarily focusing on common image processing steps used in image editing software. For cropping and resizing, the parameter is the ratio of the new area to the original area. For rotation, the parameter is the angle in degrees. For JPEG compression, the parameter is the quality factor (typically, 90\% or higher is considered high quality, 80\%-90\% is considered medium quality, and 70\%-80\% is considered low quality). For brightness, contrast, saturation, and sharpness, the parameters are the default factors used in the PIL and Torchvision libraries. For noise attack, we use uniform noise ranging from -10 to 10. For filter attack, we choose mean filtering with a 3x3 kernel. Fig.\ref{transformation} illustrates the visual effects of the same image under different transformations.

\subsection{Attack models}
To evaluate the robustness of watermarking models to intentional manipulation, We selected five state-of-the-art autoencoder models from the CompressAI library, including bmshj2018-factorized, bmshj2018-hyperprior \citep{balle2018variational}, (bmshj2018-factorized loads a factorized prior model using the ReLU activation function, while bmshj2018-hyperprior loads the original scale hyperprior model. The main difference between the two is the model structure and activation function), mbt2018, mbt2018-mean \citep{minnen2018joint} (mbt2018-mean loads a Scale Hyperprior model with non-zero mean Gaussian conditions, while mbt2018 loads a joint autoregressive hierarchical prior model. The main difference between the two is the model structure and the mean of the Gaussian conditions.), heng2020-anchor \citep{cheng2020learned} and evaluate the robustness of various watermark generation models against different types of neural autoencoder attacks at different compression rates. Generally speaking, for the same encoder, the larger the quality parameter, the higher the compression quality, the more details the compressed image retains, and the smaller the distortion, but the file size of the image will also increase accordingly.
 
\begin{figure}[h]
\centering
\includegraphics[width=1.0\columnwidth]{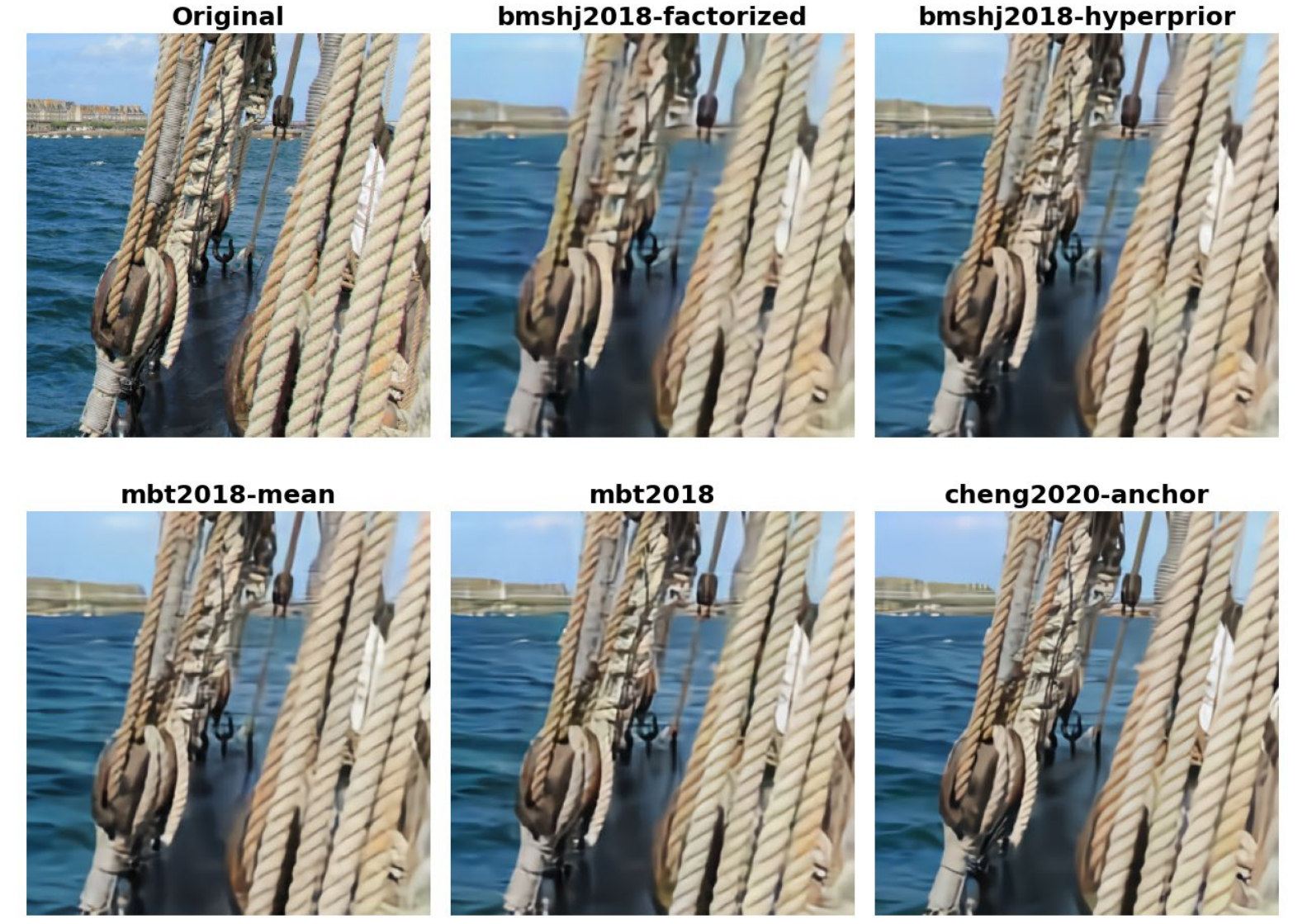}
\caption{The compression results of different encoders. }
\label{compression1}
\end{figure}

\begin{figure}[h]
\centering
\includegraphics[width=1.0\columnwidth]{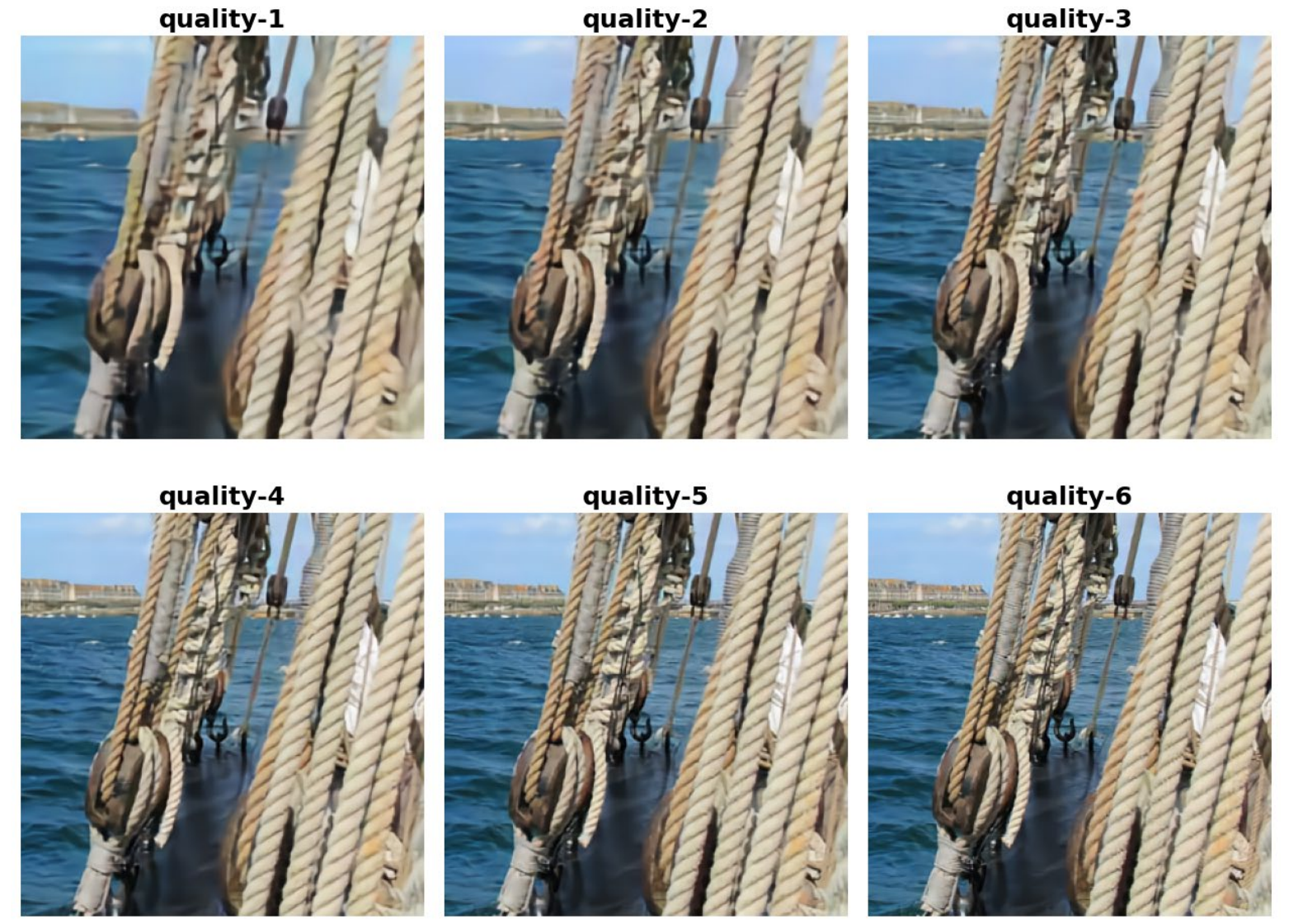}
\caption{The compression results of the same encoder (here, bmshj2018\_factorized is taken as an example) at different quality settings. For the same encoder, the higher the quality parameter, the higher the image compression quality. This means that the compressed image retains more detail and has less distortion, but the file size of the image will be correspondingly larger.}
\label{compression2}
\end{figure}

\section{Supplementary experimental results}\label{secA1}

\begin{figure*}[h]
\centering
\includegraphics[width=1.0\textwidth]{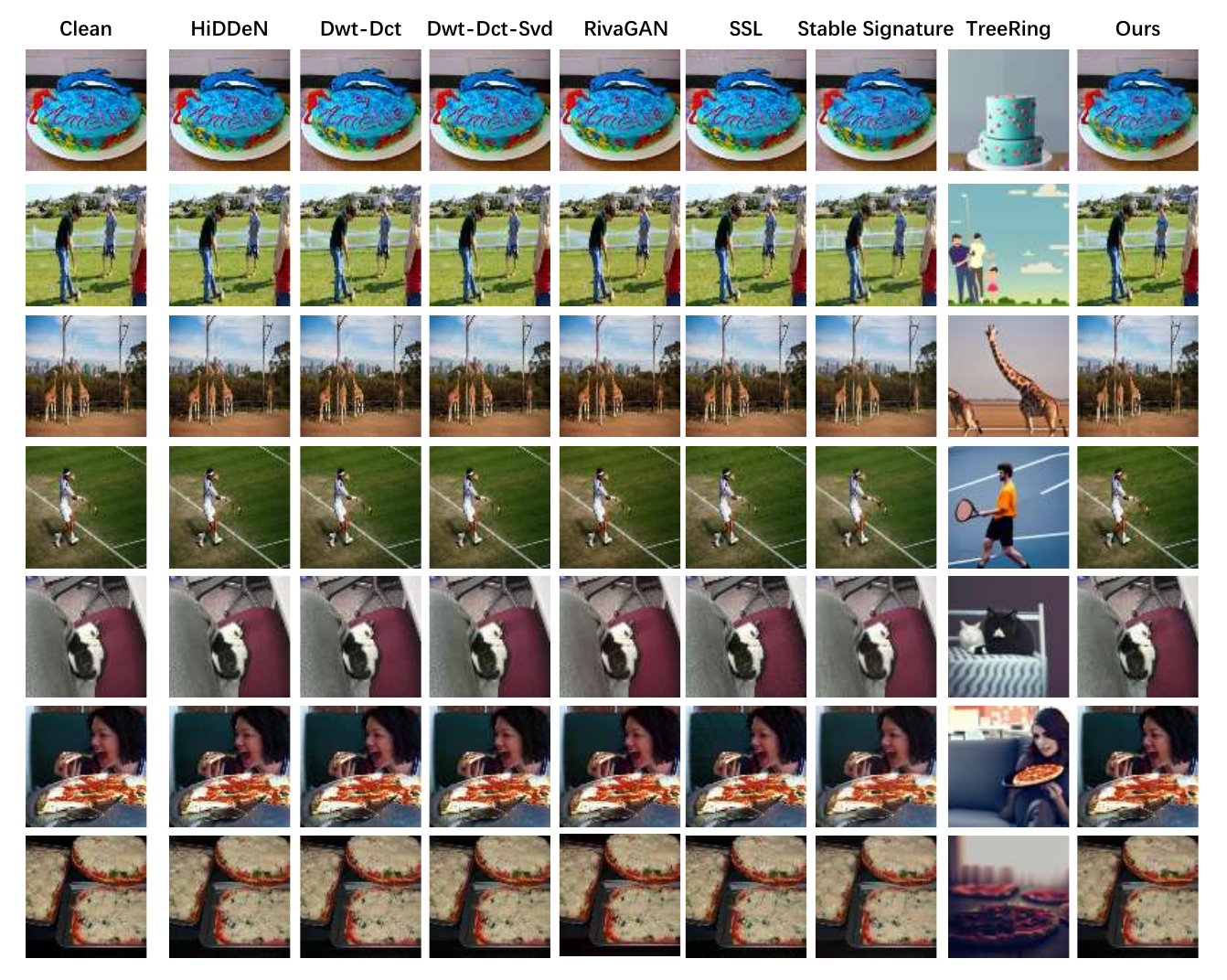}
\caption{Visual comparison of different watermarking models.}
\label{compare}
\end{figure*}

We compared the images generated by our method with those produced by other watermarking models. To ensure fairness, all methods were implemented using their original optimal settings, with the image resolution standardized to 256×256×3. During the comparison, we considered not only diffusion model-based watermarking methods (such as Stable Signature and Tree-Ring) but also watermarking models that process generated images. Specifically, these models include Dct-Dwt \citep{cox2007digital}, SSL Watermark iterative methods \citep{fernandez2022watermarking}, and encoder/decoder models similar to HiDDeN \citep{zhu2018hidden}. Notably, since the official Tree-Ring model is primarily designed for text-to-image generation, we replicated the Tree-Ring image-to-image version to maintain consistency and fairness in the comparison. Fig.\ref{compare} shows the results of the model comparisons.

Based on the experimental results, our method performs better visually compared to other diffusion model-based methods. Specifically, the watermarked image generated by the Stable Signature method is more prone to blurring, artifacts, etc., while the watermark generated by the Tree-Ring model will change the image itself, and these changes depend on the content of the prompt. Among post-processing watermarking techniques, the images generated by the SSL method exhibit obvious texture, whereas other methods show no significant difference in image quality compared to ours.

\newpage
\bibliography{aaai25}